%% file: JointGAN_camera.tex
\documentclass{article}

\usepackage{microtype}
\usepackage{graphicx}
\usepackage{subfigure}
\usepackage{booktabs} 

\usepackage{hyperref}

\usepackage[accepted]{icml2018}

\include{mlVecMat}

\icmltitlerunning{JointGAN: Multi-Domain Joint Distribution Learning with Generative Adversarial Nets}

\begin{document}

\twocolumn[
\icmltitle{JointGAN: Multi-Domain Joint Distribution Learning with \\ Generative Adversarial Nets}



\icmlsetsymbol{equal}{*}
\begin{icmlauthorlist}
	\icmlauthor{Yunchen Pu}{goo}
	\icmlauthor{Shuyang Dai}{to}
	\icmlauthor{Zhe Gan}{ed}
	\icmlauthor{Weiyao Wang}{to}
	\icmlauthor{Guoyin Wang}{to}
	\icmlauthor{Yizhe Zhang}{ed}
	\icmlauthor{Ricardo Henao}{to}
	\icmlauthor{Lawrence Carin}{to}
\end{icmlauthorlist}

\icmlaffiliation{to}{Duke University, Durham, NC, USA}
\icmlaffiliation{goo}{Facebook, Menlo Park, CA, USA}
\icmlaffiliation{ed}{Microsoft Research, Redmond, WA, USA}

\icmlcorrespondingauthor{Yunchen Pu}{pyc40@fb.com}
\icmlcorrespondingauthor{Shuyang Dai}{shuyang.dai@duke.edu}
\icmlcorrespondingauthor{Zhe Gan}{zhe.gan@microsoft.com}
\icmlcorrespondingauthor{Lawrence Carin}{lcarin@duke.edu}

\vskip 0.3in
]



\printAffiliationsAndNotice{This work was done when Yunchen Pu, Zhe Gan and Yizhe Zhang were Ph.D. students at Duke University.}  

\begin{abstract}
	A new generative adversarial network is developed for joint distribution matching.
	Distinct from most existing approaches, that only learn conditional distributions, the proposed model aims to learn a joint distribution of multiple random variables (domains).
	This is achieved by learning to sample from conditional distributions between the domains, while simultaneously learning to sample from the marginals of each individual domain.
	The proposed framework consists of multiple generators and a single softmax-based critic, all jointly trained via adversarial learning.
	From a simple noise source, the proposed framework allows synthesis of draws from the marginals, conditional draws given observations from a subset of random variables, or complete draws from the full joint distribution. Most examples considered are for joint analysis of two domains, with examples for three domains also presented.
\end{abstract}
\vspace{-4mm}
\section{Introduction}
Generative adversarial networks (GANs)~\cite{goodfellow2014generative} have emerged as a powerful framework for modeling the draw of samples from complex data distributions.
When trained on datasets of natural images, significant progress has been made on synthesizing realistic and sharp-looking images~\cite{radford2015unsupervised}.
Recent work has also extended the GAN framework for the challenging task of text generation~\cite{yu2017seqgan,zhang2017adversarial}.
However, in its standard form, GAN models distributions in one domain, \emph{i.e.}, for a single random variable.

There has been recent interest in employing GAN ideas to learn \emph{conditional} distributions for two random variables.
This setting is of interest when one desires to synthesize (or infer) one random variable given  an instance of another random variable.
Example applications include generative models with (stochastic) latent variables~\cite{mescheder2017adversarial,Tao2018chi2}, and conditional data synthesis~\cite{isola2016image,reed2016generative}, when both domains consist of observed pairs of random variables.

\begin{figure}[t!]
	\centering
	\includegraphics[width=.48\textwidth]{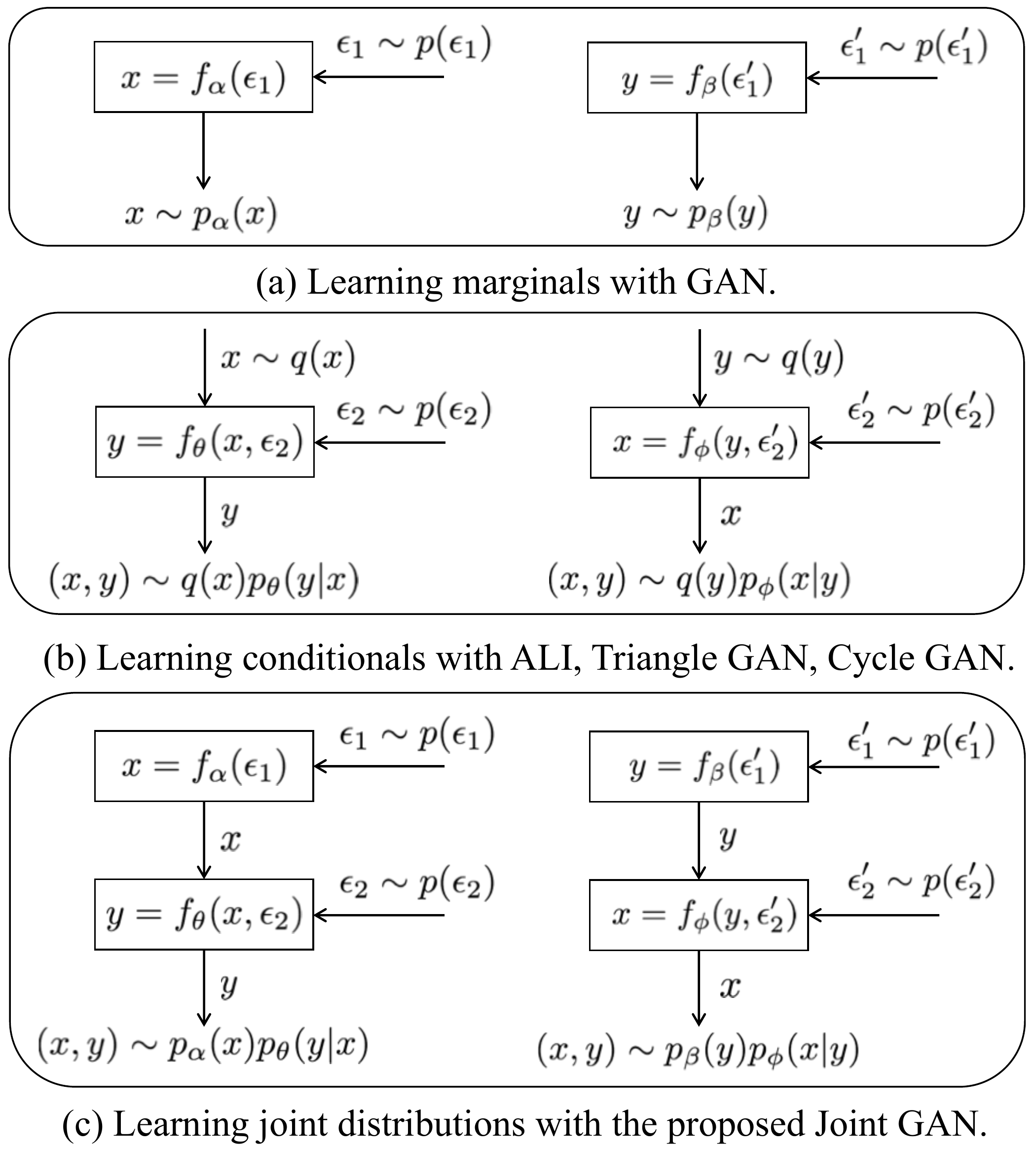}
	\vspace{-5mm}
	\caption{Comparing the generative process of GAN~\cite{goodfellow2014generative}, ALI~\cite{dumoulin2016adversarially}, Triangle GAN~\cite{gan2017triangle}, CycleGAN~\cite{zhu2017unpaired} and the proposed JointGAN model. GAN in (a) only allows synthesizing samples from marginals, the models in (b) only allow for conditional data synthesis, while the proposed model in (c), allows data synthesis from both marginals and conditionals.}
	\vspace{-1mm}
	\label{fig:generative_models}
\end{figure}

In this paper we focus on learning the {\em joint} distribution of multiple random variables using adversarial training.
For the case of two random variables, conditional GAN~\cite{mirza2014conditional} and Triangle GAN~\cite{gan2017triangle} have been utilized for this task in the case that paired data are available.
Further, adversarially learned inference (ALI)~\cite{dumoulin2016adversarially,donahue2016adversarial} and CycleGAN~\cite{zhu2017unpaired,kim2017learning,yi2017dualgan} were developed for unsupervised learning, where the two-way mappings between two domains are learned without any paired data.
These models are unified as the joint distribution matching problem by~\citet{li2017alice}.
However, in all previous approaches the joint distributions are \emph{not fully} learned, \ie, the model only learns to sample from the \emph{conditional} distributions, assuming access to the \emph{marginal} distributions, which are typically instantiated as empirical samples from each individual domain (see Figure~\ref{fig:generative_models}(b) for illustration).
Therefore, only conditional data synthesis can be achieved due to the lack of a learned sample mechanism for the marginals.

It is desirable to build a generative-model learning framework from which one may sample from a fully-learned joint distribution.
We design a new GAN framework that learns the joint distribution by decomposing it into the product of a marginal and a conditional distribution(s), each learned via adversarial training (see Figure~\ref{fig:generative_models}(c) for illustration).
The resulting model may then be employed in several distinct applications: ($i$) synthesis of draws from any of the marginals; ($ii$) synthesis of draws from the conditionals when other random variables are observed, \ie, imputation; ($iii$) or we may simultaneously draw all random variables from the joint distribution.

For the special case of two random variables, the proposed model consists of four generators and a softmax critic function.
The design includes two generators for the marginals, two for the conditionals, and a single 5-way critic (discriminator) trained to distinguish pairs of real data from four different kinds of synthetic data.
These five modules are implemented as neural networks, which share parameters for efficiency and are optimized jointly via adversarial learning. We also consider an example with three random variables.

The contributions of this work are summarized as follows.
($i$) We present the first GAN-enabled framework that allows for full joint-distribution learning of multiple random variables.
Unlike existing models, the proposed framework learns marginals and conditionals simultaneously.
($ii$) We share parameters of the generator models, and thus the resulting model does not have a significant increase in the number of parameters relative to prior work that only considered conditionals (ALI, Triangle GAN, CycleGAN, \emph{etc}.) or marginals (traditional GAN).
($iii$) Unlike existing approaches, we consolidate all real \emph{vs.} artificial sample comparisons into a single softmax-based critic function.
($iv$) While the main focus is on the case of two random variables, we extend the proposed model to learning the joint distribution of three or more random variables.
($v$) We apply the proposed model for both unsupervised and supervised learning paradigms.

\vspace{-2mm}
\section{Background}
To simplify the presentation, we first consider joint modeling of two random variables, with the setup generalized in Sec.~\ref{sec:extend} to the case of more than two domains. For the two-random-variable case, consider marginal distributions $q(\xv)$ and $q(\yv)$ defined over two random variables $\xv \in \Xcal$ and $\yv \in \Ycal$, respectively.
Typically, we have samples but not an explicit density form for $q(\xv)$ and $q(\yv)$, \emph{i.e.}, ensembles $\{\xv_i\}_{i=1}^N$ and $\{\yv_j\}_{j=1}^M$ are available for learning.
In general, their joint distribution can be written as the product of a marginal and a conditional in two ways: $q(\xv,\yv) = q(\xv) q(\yv|\xv) = q(\yv) q(\xv|\yv)$.
One random variable can be synthesized (or inferred) given the other using conditional distributions, $q(\xv|\yv)$ and $q(\yv|\xv)$.

\subsection{Generative Adversarial Networks}\label{sec:gan}
Nonparametric sampling from marginal distributions $q(\xv)$ and $q(\yv)$ can be accomplished via adversarial learning~\cite{goodfellow2014generative}, which provides a sampling mechanism that only requires gradient backpropagation, avoiding the need to explicitly adopt a form for the marginals.
Specifically, instead of sampling directly from an (assumed) parametric distribution for the desired marginal, the target random variable is generated as a deterministic transformation of an easy-to-sample, independent noise source, \emph{e.g.}, Gaussian distribution.
The sampling procedure for the marginals, \emph{implicitly} defined as $\xv \sim p_{\alphav}(\xv)$ and $\yv \sim p_{\betav} (\yv)$, is carried out through the following two generative processes:
\begin{align}
	\tilde{\xv} &= f_{\alphav} (\epsilonv_1),  \qquad \epsilonv_1 \sim p(\epsilonv_1) \,, \label{eqn:generative_x} \\
	\tilde{\yv} &= f_{\betav} (\epsilonv_1^\prime),  \qquad \epsilonv_1^\prime \sim p(\epsilonv_1^\prime) \label{eqn:generative_y} \,,
\end{align}
where $f_{\alphav}(\cdot)$ and $f_{\betav} (\cdot)$ are two \emph{marginal} generators, specified as neural networks with parameters $\alphav$ and $\betav$, respectively.
$p(\epsilonv_1)$ and $p(\epsilonv_1^\prime)$ are assumed to be simple distributions, $e.g.$, isotropic Gaussian. 
The generative processes manifested by \eqref{eqn:generative_x} and \eqref{eqn:generative_y} are illustrated in Figure~\ref{fig:generative_models}(a).
Within the models, stochasticity in $\xv$ and $\yv$ is manifested via draws from $p(\epsilonv_1)$ and $p(\epsilonv_1^\prime)$, and respective neural networks $f_{\alphav}(\cdot)$ and $f_{\betav} (\cdot)$ transform draws from these simple distributions such that they are approximately consistent with draws from $q(\xv)$ and $q(\yv)$.

For this purpose, GAN trains a $\omegav$-parameterized critic function $g_{\omegav}(\xv)$, to distinguish samples generated from $p_{\alphav}(\xv)$ and $q(\xv)$.
Formally, the minimax objective of GAN is 
\begin{align}\label{eqn:gan}
	\min_{\alphav} \max_{\omegav} \ \Lcal_{\textrm{GAN}} (\alphav,\omegav) = & \ \E_{\xv\sim q(\xv)} [\log \sigma(g_{\omegav}(\xv))] \\
	+ & \ \E_{\epsilonv_1 \sim p(\epsilonv_1)}[\log (1-\sigma(g_{\omegav}(\tilde{\xv})))] \nonumber \,,
\end{align}
with expectations $\E_{\xv\sim q(\xv)}[\cdot]$ and $\E_{\epsilonv_1\sim p(\epsilonv_1)}[\cdot]$ approximated via sampling, and $\sigma(\cdot)$ is the sigmoid function. As shown in~\citet{goodfellow2014generative}, the equilibrium of the objective in \eqref{eqn:gan} is achieved if and only if $p_{\alphav}(\xv) = q(\xv)$.
%

Similarly, we can design a corresponding minimax objective that is similar to \eqref{eqn:gan} to match the marginal $p_{\betav}(\yv)$ to $q(\yv)$.
\vspace{-2mm}
\subsection{Adversarially Learned Inference}\label{sec:ali}
In the same spirit, sampling from conditional distributions $q(\xv|\yv)$ and $q(\yv|\xv)$ can be also achieved as a deterministic transformation of two inputs, the variable in the source domain as a \emph{covariate}, plus an independent source of noise.
Specifically, the sampling procedure for the conditionals $\yv\sim p_{\thetav}(\yv|\xv)$ and $\xv\sim p_{\phiv}(\xv|\yv)$ is modeled as 
\begin{align}
	\tilde{\yv} &= f_{\thetav} (\xv,\epsilonv_2), \quad \xv\sim q(\xv), \quad \epsilonv_2 \sim p(\epsilonv_2) \,, \label{eqn:ali_x} \\
	\tilde{\xv} &= f_{\phiv} (\yv,\epsilonv_2^\prime), \quad \yv\sim q(\yv), \quad \epsilonv_2^\prime \sim p(\epsilonv_2^\prime) \,, \label{eqn:ali_y}
\end{align}
where $f_{\thetav}(\cdot)$ and $f_{\phiv}(\cdot)$ are two \emph{conditional} generators, specified as neural networks with parameters $\thetav$ and $\phiv$, respectively.
In practice, the inputs of $f_{\thetav}(\cdot)$ and $f_{\phiv}(\cdot)$ are concatenated.
As in GAN, $p(\epsilonv_2)$ and $p(\epsilonv_2^\prime)$ are two simple distributions that provide the stochasticity when generating $\yv$ given $\xv$, and \emph{vice versa}.
The conditional generative processes manifested in \eqref{eqn:ali_x} and \eqref{eqn:ali_y} are illustrated in Figure~\ref{fig:generative_models}(b),

ALI~\cite{dumoulin2016adversarially} learns the two desired conditionals by matching joint distributions $p_{\thetav} (\xv,\yv) = q(\xv)p_{\thetav}(\yv|\xv)$ and $p_{\phiv} (\xv,\yv)= q(\yv)p_{\phiv}(\xv|\yv)$, using a critic function $g_{\omegav}(\xv,\yv)$ similar to \eqref{eqn:gan}.
The minimax objective of ALI can be written as
\begin{align}
	\min_{\thetav,\phiv} \max_{\omegav} \quad & \ \E_{(\xv,\tilde{\yv})\sim p_{\thetav}(\xv,\yv)} [\log \sigma(g_{\omegav}(\xv,\tilde{\yv}))] \label{eqn:ali} \\
	+ & \ \E_{(\tilde{\xv},\yv)\sim p_{\phiv}(\xv,\yv)} [\log (1-\sigma(g_{\omegav}(\tilde{\xv},\yv)))]\,. \nonumber 
\end{align}
The equilibrium of the objective in (\ref{eqn:ali}) is achieved if and only if $p_{\thetav}(\xv,\yv) = p_{\phiv} (\xv,\yv)$.
%


While ALI is able to match joint distributions using \eqref{eqn:ali}, only conditional distributions $p_{\thetav}(\yv|\xv)$ and $p_{\phiv}(\xv|\yv)$ are learned, thus assuming access to (samples from) the true marginal distributions $q(\xv)$ and $q(\yv)$, respectively.
\section{Adversarial Joint Distribution Learning}\label{sec:model}
Below we discuss \emph{fully} learning the joint distribution of two random variables in both supervised and unsupervised settings. By ``supervised'' it is meant that we have access to joint empirical data $(\xv,\yv)$, and by ``unsupervised'' it is meant that we have access to empirical draws of $\xv$ and $\yv$, but not paired observations $(\xv,\yv)$ from the joint distribution.
\subsection{JointGAN} \label{sec:joint_gan}
Since $q(\xv,\yv) = q(\xv) q(\yv|\xv) = q(\yv)q(\xv|\yv)$, a simple way to achieve joint-distribution learning is to first learn models for the two marginals separately, using a pair of traditional GANs, followed by training an independent ALI model for the two conditionals.
However, such a two-step training procedure is suboptimal, as there is no information flow between marginals and conditionals during training.
This suboptimality is demonstrated in the experiments.
Additionally, a two-step learning process becomes cumbersome when considering more than two random variables.

Alternatively, we consider learning to sample from conditionals via $p_{\thetav}(\yv|\xv)$ and $p_{\phiv}(\xv|\yv)$, while also learning to sample from marginals via $p_{\alphav}(\xv)$ and $p_{\betav}(\yv)$. All model training is performed simultaneously. We term our model \emph{JointGAN} for full GAN analysis of joint random variables.

\paragraph{Access to Paired Empirical Draws} In this setting, we assume access to samples from the true (empirical) joint distribution $q(\xv,\yv)$.
%
The models we seek to learn constitute two means of approximating draws from the true distribution $q(\xv,\yv)$, \emph{i.e.}, $p_{\alphav}(\xv) p_{\thetav}(\yv|\xv)$ and $p_{\betav}(\yv) p_{\phiv}(\xv|\yv)$, as shown in Figure~\ref{fig:generative_models}(c): 
\begin{align}
	\tilde{\xv} &= f_{\alphav} (\epsilonv_1), \qquad \tilde{\yv} = f_{\thetav} (\tilde{\xv},\epsilonv_2),  \label{eqn:hierGAN_x} \\
	\hat{\yv} &= f_{\betav} (\epsilonv_1^\prime), \qquad \hat{\xv} = f_{\phiv} (\hat{\yv},\epsilonv_2^\prime), \label{eqn:hierGAN_y}
\end{align}
where $f_{\alphav}(\cdot)$, $f_{\betav}(\cdot)$, $f_{\thetav}(\cdot)$ and $f_{\phiv}(\cdot)$ are neural networks as defined previously.
$\epsilonv_1$, $\epsilonv_2$, $\epsilonv_1^\prime$ and $\epsilonv_2^\prime$ are independent noise.
Note that the only difference between $f_{\alphav}(\cdot)$ and $f_{\phiv}(\cdot)$ is that the function $f_{\phiv}(\cdot)$ has another conditional input $\yv$ when compared with $f_{\alphav}(\cdot)$.
Therefore, in implementation, we couple the parameters $\alphav$ and $\phiv$ together. Similarly, $\betav$ and $\thetav$ are also coupled together. 
Specifically, $f_\alphav(\cdot)$ and $f_\betav(\cdot)$ are implemented as
\begin{align}
f_\alphav(\cdot) = f_\phiv(\zerov,\, \cdot),\qquad f_\betav(\cdot) = f_\thetav(\zerov,\cdot) \,,
\end{align}
where $\zerov$ is an all-zero tensor which has the same size as $\xv$ or $\yv$.
As a result, \eqref{eqn:hierGAN_x} and \eqref{eqn:hierGAN_y} have almost the same number of parameters as ALI-like approaches.


The following notation is introduced for simplicity of illustration:
\begin{align}
	p_1(\xv,\yv) & = q(\xv) p_{\thetav}(\yv|\xv), & p_2(\xv,\yv) & = q(\yv) p_{\phiv}(\xv|\yv) \nonumber \\
	p_3(\xv,\yv) & = p_{\alphav}(\xv) p_{\thetav}(\yv|\xv), & p_4(\xv,\yv) & = p_{\betav}(\yv) p_{\phiv}(\xv|\yv) \nonumber \\
	p_5 (\xv,\yv) & = q(\xv,\yv) \,, \label{eqn:five_dists}
\end{align}
where $p_1(\xv,\yv)$, $p_2(\xv,\yv)$, $p_3(\xv,\yv)$ and $p_4(\xv,\yv)$ are implicitly defined in \eqref{eqn:ali_x}, \eqref{eqn:ali_y}, \eqref{eqn:hierGAN_x} and \eqref{eqn:hierGAN_y}. Note that $p_5 (\xv,\yv)$ is simply the empirical joint distribution.

When learning, we wish to impose that the five distributions in \eqref{eqn:five_dists} should be identical.
Toward this end, an adversarial objective is specified.
Joint pairs $(\xv,\yv)$ are drawn from the five distributions in \eqref{eqn:five_dists}, and a critic function is learned to discriminate among them, while the four generators are trained to mislead the critic.
Naively, for JointGAN, one can use 4 binary critics to mimic a 5-class classifier.
Departing from previous work such as~\citet{gan2017triangle}, here the discriminator is implemented directly as a 5-way softmax classifier.
Compared with using multiple binary classifiers, this design is more principled in that we avoid multiple critics resulting in possibly conflicting (real \emph{vs.} synthetic) assessments.

Let the critic $g_{\omegav}(\xv , \yv)\in \Delta^{4}$ (in the 4-simplex) be a $\omegav$-parameterized neural network with softmax on the top layer, \ie, $\sum_{k=1}^{5}g_{\omegav}(\xv, \yv)[k]=1$ and $g_{\omegav}(\xv, \yv)[k]\in (0, 1)$, where $g_{\omegav}(\xv, \yv)[k]$ is an entry of $g_{\omegav}(\xv , \yv)$.
The minimax objective for JointGAN, $\Lcal_{\textrm{JointGAN}} (\thetav,\phiv, \omegav)$, is given by
%
%
%
\begin{align}
	& \min_{\thetav,\phiv} \max_{\omegav} \ \Lcal_{\textrm{JointGAN}} (\thetav,\phiv, \omegav) \label{eqn:proposed_gan} \\
	& \hspace{30mm} = \textstyle{\sum_{k=1}^5} \E_{p_k(\xv,\yv)} [\log g_{\omegav}(\xv,\yv)[k]] \,. \nonumber
\end{align}
The above objective \eqref{eqn:proposed_gan} has taken into consideration the model design that $\alphav$ and $\phiv$ are coupled together, with the same for $\betav$ and $\thetav$; thus $\alphav$ and $\betav$ are not present in \eqref{eqn:proposed_gan}.
Note that expectation $\E_{p_5(\xv,\yv)}[\cdot]$ is approximated using \emph{empirical} joint samples, expectations $\E_{p_3(\xv,\yv)}[\cdot]$ and $\E_{p_4(\xv,\yv)}[\cdot]$ are both approximated with \emph{purely synthesized} joint samples, while $\E_{p_1(\xv,\yv)}[\cdot]$ and $\E_{p_2(\xv,\yv)}[\cdot]$ are approximated using \emph{conditionally synthesized} samples, given samples from the empirical marginals.
The following proposition characterizes the solution of \eqref{eqn:proposed_gan} in terms of the joint distributions.
\begin{prop}
	The equilibrium for the minimax objective in \eqref{eqn:proposed_gan} is achieved if and only if $p_1(\xv,\yv) = p_2(\xv,\yv) = p_3(\xv,\yv) = p_4(\xv,\yv) = p_5(\xv,\yv)$.
\end{prop}
The proof is provided in Appendix A.

\paragraph{No Access to Paired Empirical Draws} When paired data samples are not available, we do not have access to draws from $p_5(\xv,\yv)=q(\xv,\yv)$, so this term is not considered in \eqref{eqn:proposed_gan}.
Instead, we wish to impose ``cycle consistency''~\cite{zhu2017unpaired}, \emph{i.e.}, $q(\xv)\rightarrow \xv \rightarrow p_{\thetav}(\yv|\xv)\rightarrow \yv \rightarrow p_{\phiv}(\xv|\yv) \rightarrow \hat{\xv}$ yields small $||\xv-\hat{\xv}||$, for an appropriate norm.
Similarly, we impose $q(\yv)\rightarrow \yv \rightarrow p_{\phiv}(\xv|\yv)\rightarrow \xv \rightarrow p_{\thetav}(\yv|\xv) \rightarrow \hat{\yv}$ resulting in small $||\yv -\hat{\yv}||$.

In this case, the discriminator becomes a 4-class classifier.
Let $g^\prime_{\omegav}(\xv,\yv)\in\Delta^3$ denote a new critic, with softmax on the top layer, \ie, $\sum_{k=1}^{4}g^\prime_{\omegav}(\xv, \yv)[k]=1$ and $g^\prime_{\omegav}(\xv, \yv)[k]\in (0, 1)$.
To encourage cycle consistency, we modify the objective in \eqref{eqn:proposed_gan} as
%
%
%
\begin{align}
	& \min_{\thetav,\phiv} \max_{\omegav} \ \Lcal_{\textrm{JointGAN}} (\thetav,\phiv, \omegav) \label{eqn:proposed_gan_unsup} \\
	& \hspace{11mm} = \textstyle{\sum_{k=1}^4} \E_{p_k(\xv,\yv)} [\log g^\prime_{\omegav}(\xv,\yv)[k]] + R_{\thetav,\phiv}(\xv,\yv) \,, \nonumber
\end{align}
where $R_{\thetav,\phiv}(\xv, \yv)$ in \eqref{eqn:proposed_gan_unsup} is a cycle-consistency regularization term specified as
\begin{align*}
	R_{\thetav,\phiv}(\xv, \yv) = & \ \E_{\xv\sim q(\xv), \yv\sim p_{\thetav}(\yv|\xv), \hat{\xv}\sim p_{\phiv}(\xv|\yv)} ||\xv-\hat{\xv}|| \\
	+ & \ \E_{\yv\sim q(\yv), \xv\sim p_{\phiv}(\xv|\yv), \hat{\yv}\sim p_{\thetav}(\yv|\xv)}||\yv-\hat{\yv}|| \,.
\end{align*}

\subsection{Extension to multiple domains\label{sec:extend}}

\begin{figure}[t!]
	\centering
	\includegraphics[width=.3\textwidth]{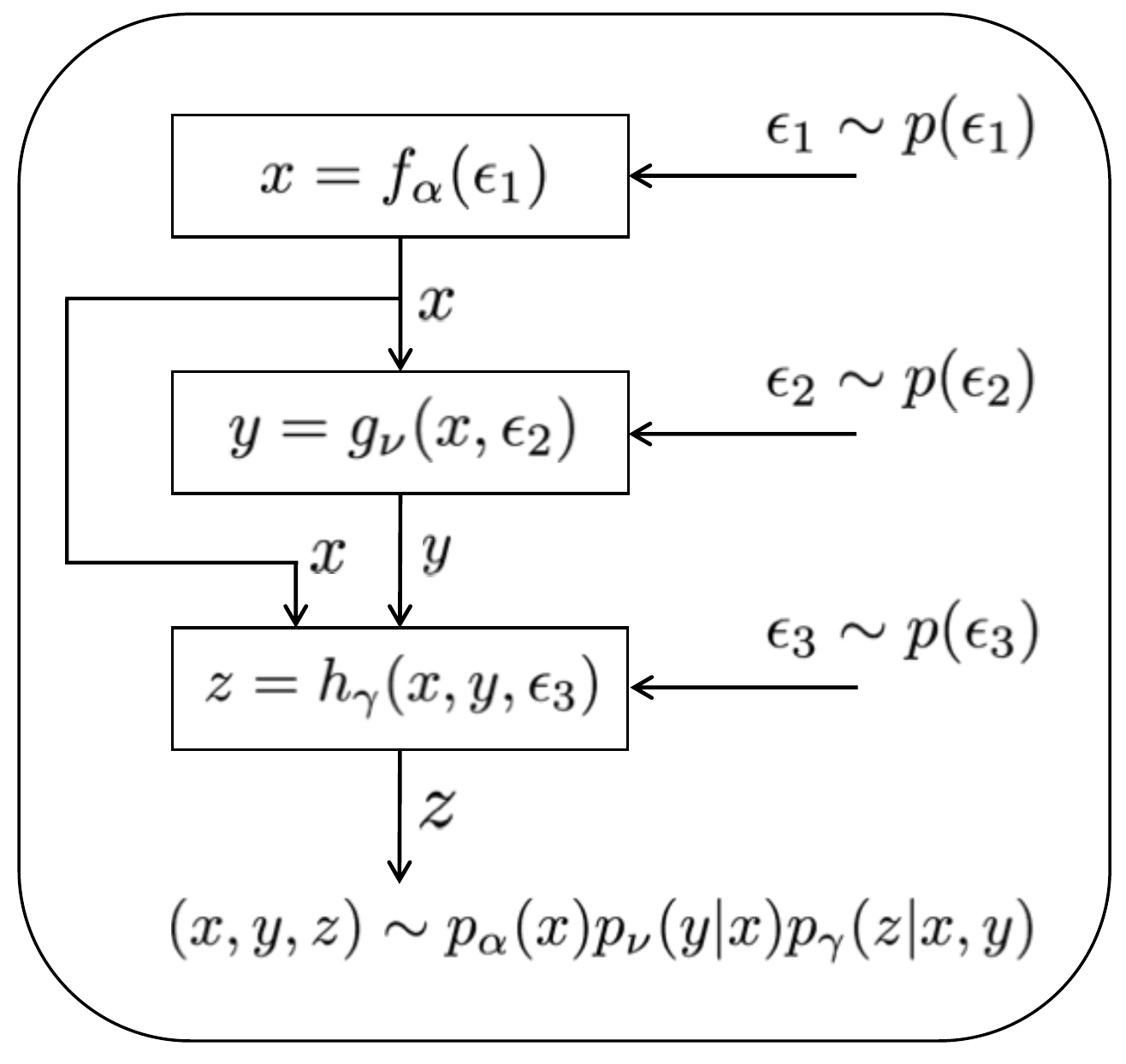}
	\vspace{-2mm}
	\caption{Generative model for the tuple $(\xv,\yv,\zv)$ manifested via $p_{\alphav}(\xv)p_{\nuv}(\yv|\xv)p_{\gammav}(\zv|\xv,\yv)$. Note the skip connection that naturally arises due to the form of the generative model.
	}
	\vspace{-2mm}
	\label{fig:model_for_three_domains}
\end{figure}
%
%
%
%
The above formulation may be extended to the case of three or more joint random variables.
However, for $m$ random variables, there are $m!$ different ways in which the joint distribution can be factorized.
For example, for joint random variables $(\xv,\yv,\zv)$, there are possibly six different forms of the model.
One must have access to all the six instantiations of these models, if the goal is to be able to generate (impute) samples from all conditionals. However, not all modeled forms of $p(\xv,\yv,\zv)$ need to be considered, if there is not interest in the corresponding form of the conditional. Below, we consider two specific forms of the model:
\begin{align}
	p(\xv,\yv,\zv)&=p_{\alphav}(\xv)p_{\nuv}(\yv|\xv)p_{\gammav}(\zv|\xv,\yv)\\
	&=p_{\betav}(\zv)p_{\psiv}(\yv|\zv)p_{\etav}(\xv|\yv,\zv) \,.
\end{align}
Typically, the joint draws from $q(\xv,\yv,\zv)$ may not be easy to access; therefore, we assume that only empirical draws from $q(\xv,\yv)$ and $q(\yv,\zv)$ are available. For the purpose of adversarial learning, we let the critic be a 6-class softmax classifier that aims to distinguish samples from the following 6 distributions: 
\begin{align*}
	p_{\alphav}(\xv) p_{\nuv}(\yv|\xv) p_{\gammav}(\zv|\xv,\yv)\,, & \qquad p_{\betav}(\zv) p_{\psiv}(\zv|\yv) p_{\etav}(\xv|\yv,\zv) \\
	q(\xv) p_{\nuv}(\yv|\xv) p_{\gammav}(\zv|\xv,\yv)\,, & \qquad q(\xv) p_{\psiv}(\zv|\yv) p_{\etav}(\xv|\yv,\zv) \\
	q(\xv,\yv) p_{\gammav}(\zv|\xv,\yv)\,, & \qquad q(\yv,\zv) p_{\etav}(\xv|\yv,\zv) \,.
\end{align*}
After training, one may synthesize $(\xv,\yv,\zv)$, impute $(\yv,\zv)$ from observed $\xv$, or impute $\zv$ from $(\xv,\yv)$, \emph{etc.}
Examples of this learning paradigm is demonstrated in the experiments. 
Interestingly, when implementing a sampling-based method for the above models, skip connections are manifested naturally as a result of the partitioning of the joint distribution, \eg, via $p_{\gammav}(\zv|\xv,\yv)$ and $p_{\etav}(\xv|\yv,\zv)$.
This is illustrated in Figure \ref{fig:model_for_three_domains}, for $p_{\alphav}(\xv)p_{\nuv}(\yv|\xv)p_{\gammav}(\zv|\xv,\yv)$.
	
\vspace{-2mm}
\section{Related work}
\vspace{-1mm}
Adversarial methods for joint distribution learning can be roughly divided into two categories, depending on the application:
($i$) \emph{generation and inference} if one of the domains consists of (stochastic) latent variables, and
($ii$) \emph{conditional data synthesis} if both domains consists of observed pairs of random variables.
Below, we review related work from these two perspectives.

\vspace{-3mm}
\paragraph{Generation and inference}
The joint distribution of data and latent variables or \emph{codes} can be considered in two (symmetric) forms: ($i$) from observed data samples fed through the encoder to yield codes, \ie, \emph{inference}, and ($ii$) from codes drawn from a simple prior and propagated through the decoder to manifest data samples, \ie, \emph{generation}.
ALI~\cite{dumoulin2016adversarially} and BiGAN~\cite{donahue2016adversarial} proposed fully adversarial methods for this purpose.
There are also many recent works concerned with integrating variational autoencoder (VAE)~\cite{kingma2013auto,pu2016variational} and GAN concepts for improved data generation and latent code inference~\cite{hu2017unifying}.
Representative work includes the AAE~\cite{makhzani2015adversarial}, VAE-GAN~\cite{larsen2015autoencoding}, AVB~\cite{mescheder2017adversarial}, AS-VAE~\cite{pu2017adversarial}, SVAE~\cite{pu2017symmetric}, \emph{etc}.

\vspace{-3mm}
\paragraph{Conditional data synthesis}
Conditional GAN can be readily used for conditional-data synthesis if paired data are available.
Multiple conditional GANs have been proposed to generate the images based on class labels~\cite{mirza2014conditional}, attributes~\cite{perarnau2016invertible}, text~\cite{reed2016generative,xu2017attngan} 
and other images~\cite{isola2016image}.
Often, only the mapping from one direction (a single conditional) is learned.
Triangle GAN~\cite{gan2017triangle} and Triple GAN~\cite{chongxuan2017triple} can be used to learn bi-directional mappings (both conditionals) in a semi-supervised learning setup. 
Unsupervised learning methods were also developed for this task.
CycleGAN~\cite{zhu2017unpaired} proposed to use two generators to model the conditionals and two critics to decide whether a generated sample is synthesized, in each individual domain.
Further, additional reconstruction losses were introduced to impose cycle consistency.
Similar work includes DiscoGAN~\cite{kim2017learning}, DualGAN~\cite{yi2017dualgan} and UNIT~\cite{liu2017unsupervised}.

CoGAN~\cite{liu2016coupled} can be used to achieve joint distribution learning. However, the joint distribution is only roughly approximated by the marginals, via sharing low-layer weights of the generators, hence not learning the true (empirical) joint distributions in a principled way.

All the other previously proposed models focus on learning to sample from the conditionals given samples from one of the true (empirical) marginals, while the proposed model, to the best of the authors' knowledge, is the first attempt to learn a full joint distribution of two or more observed random variables. Moreover, this paper presents the first consolidation of multiple binary critics into a single unified softmax-based critic. 

We observe that the proposed model, JointGAN, may follow naturally in concept from GAN~\cite{goodfellow2014generative} and ALI~\cite{donahue2016adversarial,dumoulin2016adversarially}. However, there are several keys to obtaining good performance. Specifically, ($i$) the condition distribution setup naturally yields skip connections in the architecture. ($ii$) Compared with using multiple binary critics, the softmax-based critic can be considered as sharing the parameters among all the binary critics except the top layer. This also imposes the critic embedding the generated samples from different ways into a common latent space and reduces the number of parameters. ($iii$) The weight-sharing constraint among generators enforces that synthesized images from the marginal and conditional generator share a common latent space, and also further reduces the number of parameters in the network.

\vspace{-2mm}
\section{Experiments\label{sec:exp}}
Adam \cite{kingma2014adam} with learning rate 0.0002 is utilized for optimization of the JointGAN objectives. All noise vectors $\epsilonv_1$, $\epsilonv_2$, $\epsilonv^\prime_1$ and $\epsilonv^\prime_2$ are drawn from a $\Ncal(0,\Imat)$ distribution, with the dimension of each set to 100. Besides the results presented in this section, more results can be found in Appendix C.2. The code can be found at \url{https://github.com/sdai654416/Joint-GAN}.

\begin{figure*}[tbh!]
	\centering
	\includegraphics[width=0.9\linewidth]{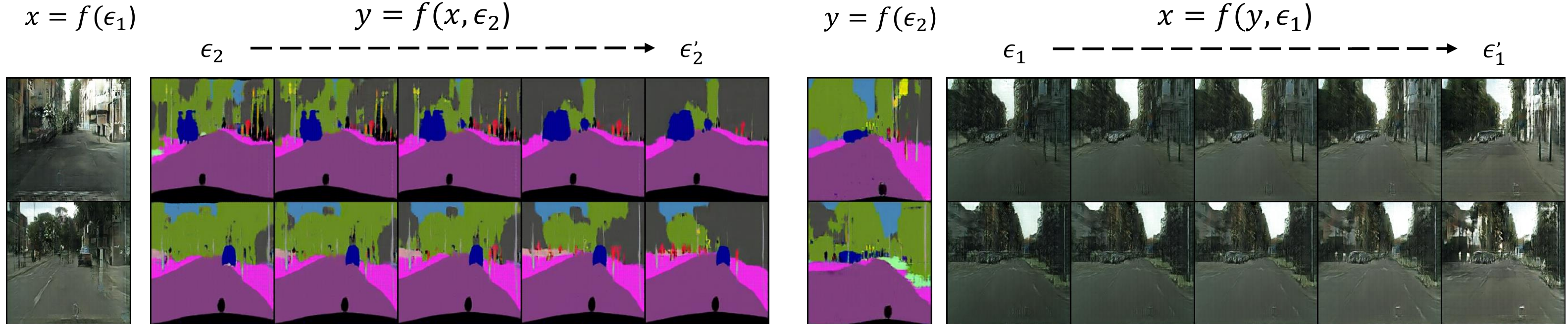}\\
	\includegraphics[width=0.9\linewidth]{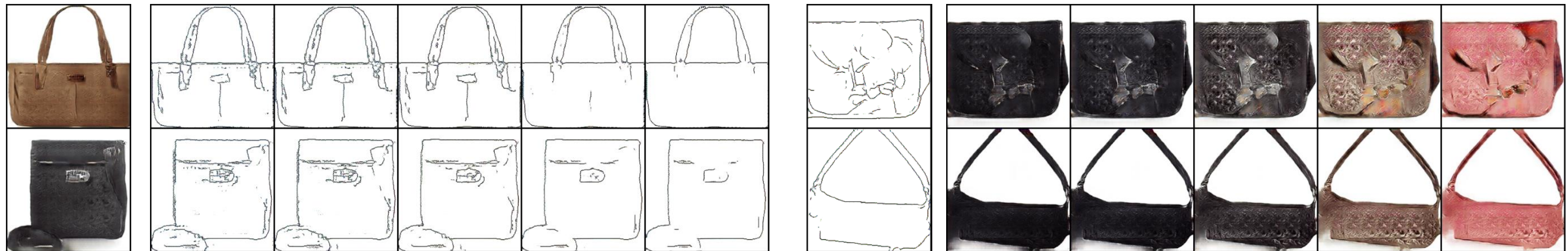}\\
	\includegraphics[width=0.9\linewidth]{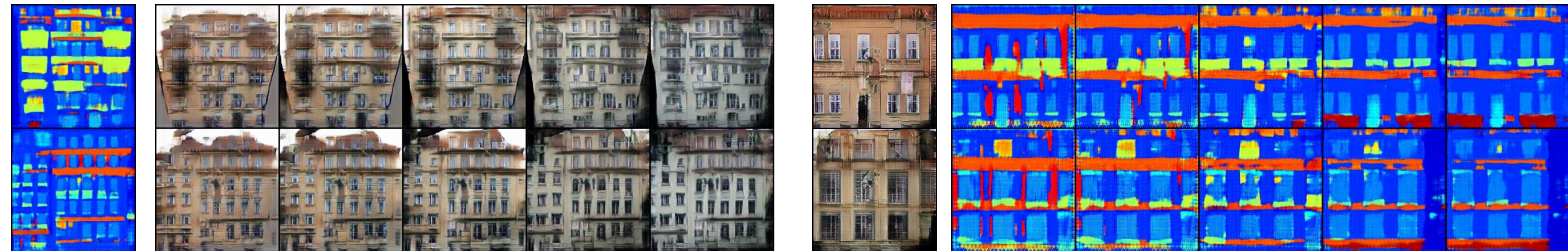}
	\vspace{-2mm}
	\caption{\small{Generated paired samples from models trained on paired data.}}\label{fig:more_paired}
	\vspace{-3mm}
\end{figure*}
\vspace{-1mm}
\subsection{Joint modeling multi-domain images}
\paragraph{Datasets}
We present results on five datasets: edges$\leftrightarrow$ shoes~\cite{yu2014fine}, edges$\leftrightarrow$handbags~\cite{zhu2016generative}, Google maps$\leftrightarrow$aerial photos~\cite{isola2016image}, labels$\leftrightarrow$facades~\cite{tylevcek2013spatial} and labels $\leftrightarrow$cityscapes~\cite{cordts2016cityscapes}. All of these datasets are two-domain image pairs. 

For three-domain images, we create a new dataset by combining labels$\leftrightarrow$facades pairs and labels$\leftrightarrow$cityscapes pairs into facades$\leftrightarrow$labels$\leftrightarrow$cityscapes tuples. In this dataset, only empirical draws from $q(\xv,\yv)$ and $q(\yv,\zv)$ are available. Another new dataset is created based on MNIST, where the three image domains are the MNIST images, clockwise transposed ones, and anticlockwise transposed ones. 

\vspace{-3mm}
\paragraph{Baseline}
As described in Sec.~\ref{sec:joint_gan}, a two-step model is implemented as the baseline. Specifically, WGAN-GP~\citep{iwgan} is employed to model the two marginals; Pix2pix~\citep{isola2016image} and CycleGAN~\citep{zhu2017unpaired} are utilized to model the conditionals for the case with and without access to paired empirical draws, respectively.

\vspace{-3mm}
\paragraph{Network Architectures}
For generators, we employed the U-net~\citep{Unet} which has been demonstrated to achieve impressive results for image-to-image translation. 
Following~\citet{isola2016image}, PatchGAN is employed for the discriminator, which provides real \emph{vs.} synthesized prediction on $70\times 70$ overlapping image patches.

\subsubsection{Qualitative Results}
\begin{figure}[tb!]
	\centering
	\includegraphics[width=0.99\linewidth]{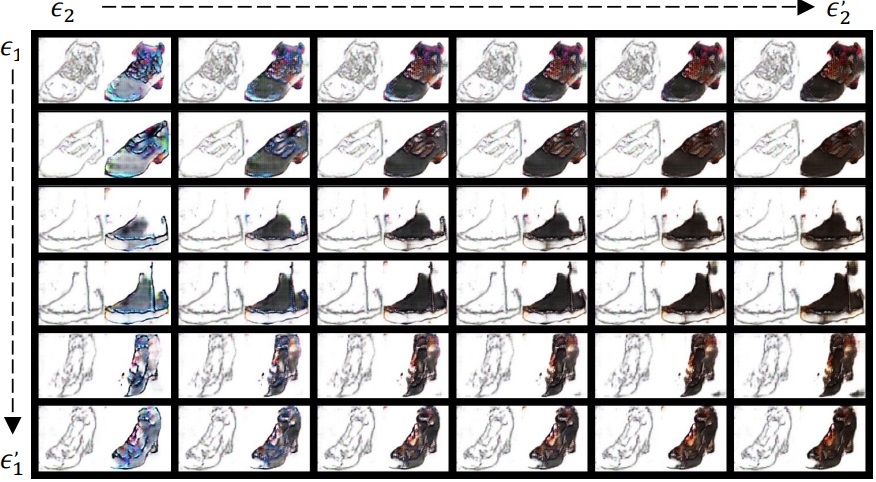}
	\vspace{-2mm}
	\caption{\small{Generated paired samples from the JointGAN model trained on the paired edges$\leftrightarrow$shoes dataset.}}\label{fig:shoes_paired}
	\vspace{-3mm}
\end{figure}

Figures~\ref{fig:more_paired} and~\ref{fig:shoes_paired} show the results trained on paired data. All the image pairs are generated from random noise. For Figure~\ref{fig:shoes_paired}, we first draw $(\epsilonv_1, \epsilonv_2)$ and $(\epsilonv^\prime_1, \epsilonv^\prime_2)$ to generate the top-left image pairs and bottom-right image pairs according to (\ref{eqn:hierGAN_x}). All remaining image pairs are generated from the noise pair made by linear interpolation between $\epsilonv_1$ and $\epsilonv^\prime_1$, and between $\epsilonv_2$ and $\epsilonv^\prime_2$, respectively, also via (\ref{eqn:hierGAN_x}).  
For Figure~\ref{fig:more_paired}, in each row of the left block, the column is first generated from $p_{\alphav}(\xv)$, and then the images of the right part are generated based on the leftmost image and an additional noise vector linear-interpolated between two random points $\epsilonv_2$ and $\epsilonv^\prime_2$.
The images in the right block are produced in a similar way. 

\begin{figure}[tb!]
	\centering
	\includegraphics[width=0.95\linewidth]{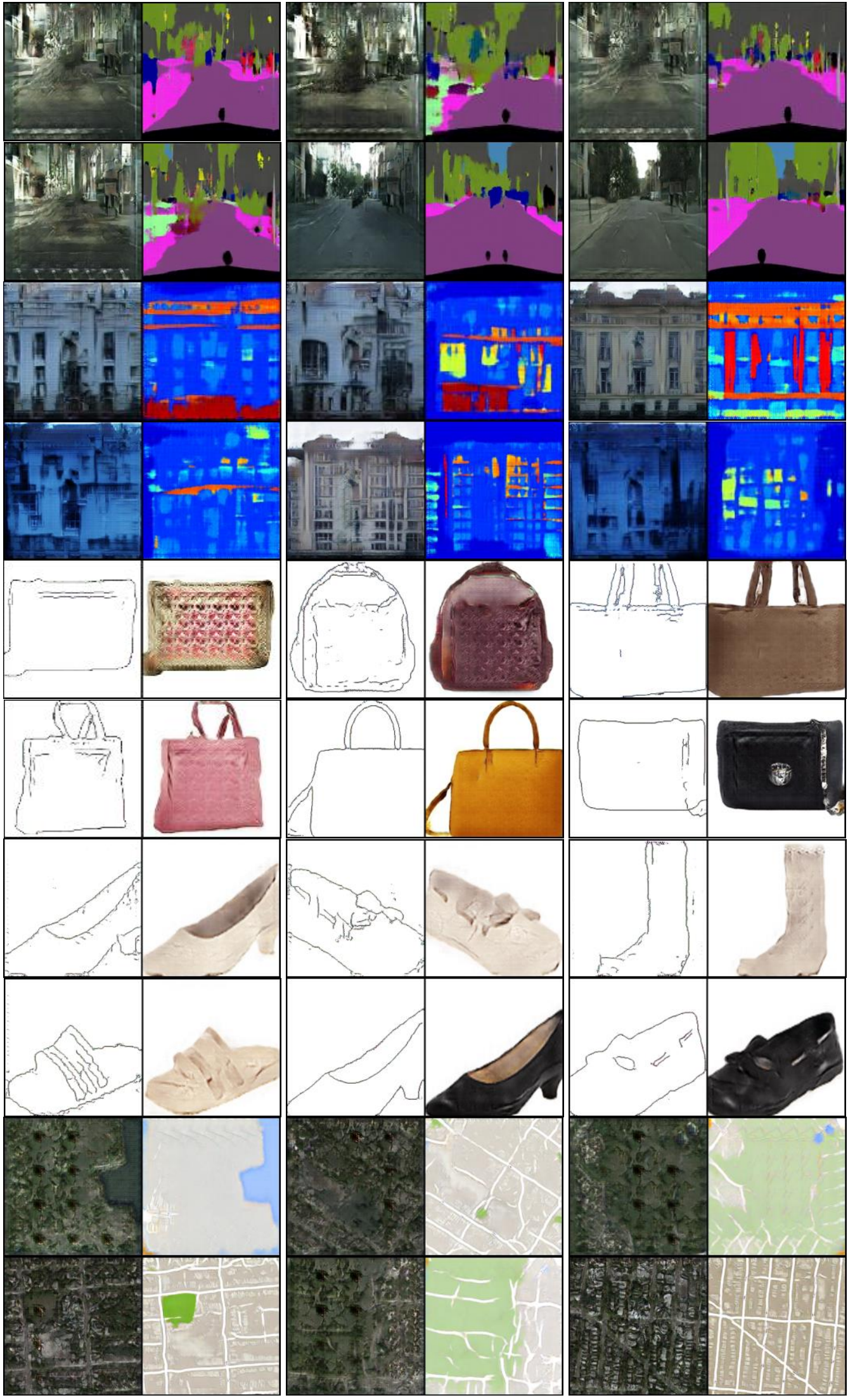}
	\vspace{-2mm}
	\caption{\small{Generated paired samples from models trained on unpaired data.}}\label{fig:unpaired}
	\vspace{-3mm}
\end{figure}

\begin{figure}[t!]
	\centering
	\includegraphics[width=0.95\linewidth]{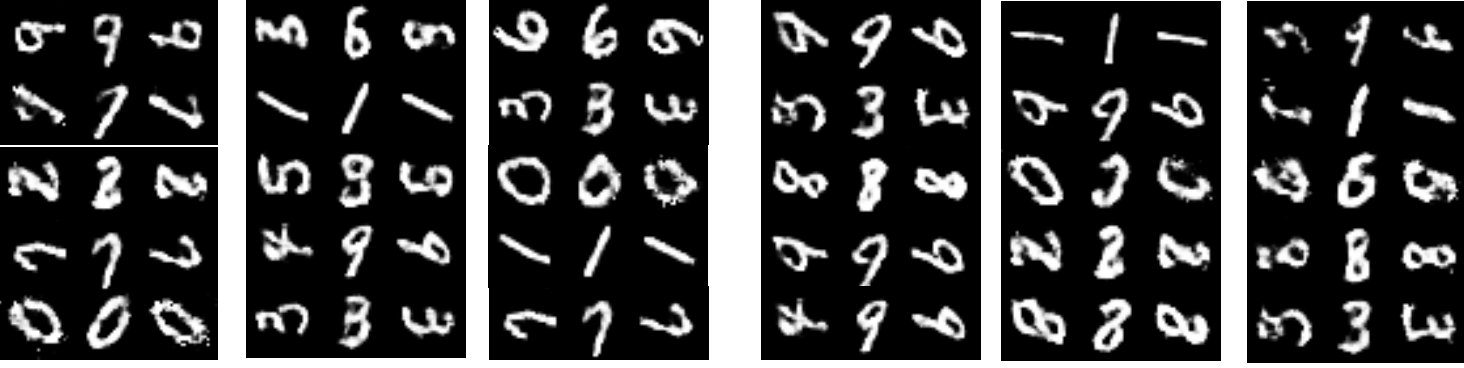}
	\vspace{-2mm}
	\caption{\small{Generated three-domain samples from models trained on MNIST. The images in each tuple of the left three columns are sequentially generated from left to right, while those in the right three columns are generated from right to left. }}\label{fig:three_mnist}
	\vspace{-3mm}
\end{figure}

These results demonstrate that our model is able to generate both realistic and highly coherent image pairs. In addition, the interpolation experiments illustrate that our model maintains smooth transitions in the latent space, with each point in the latent space corresponding to a plausible image. For example, in the edges$\leftrightarrow$handbags dataset, it can be seen that the edges smoothly transforming from complicated structures into simple ones, and the color of the handbags transforming from black to red. 
The quality of images generated from the baseline is much worse than ours, and are provided in Appendix C.1.

Figure~\ref{fig:unpaired} shows the generated samples trained on unpaired data. Our model is able to produce image pairs whose quality are close to the samples trained on paired data.

\begin{figure*}[tb!]
	\centering
	\includegraphics[width=0.9\linewidth]{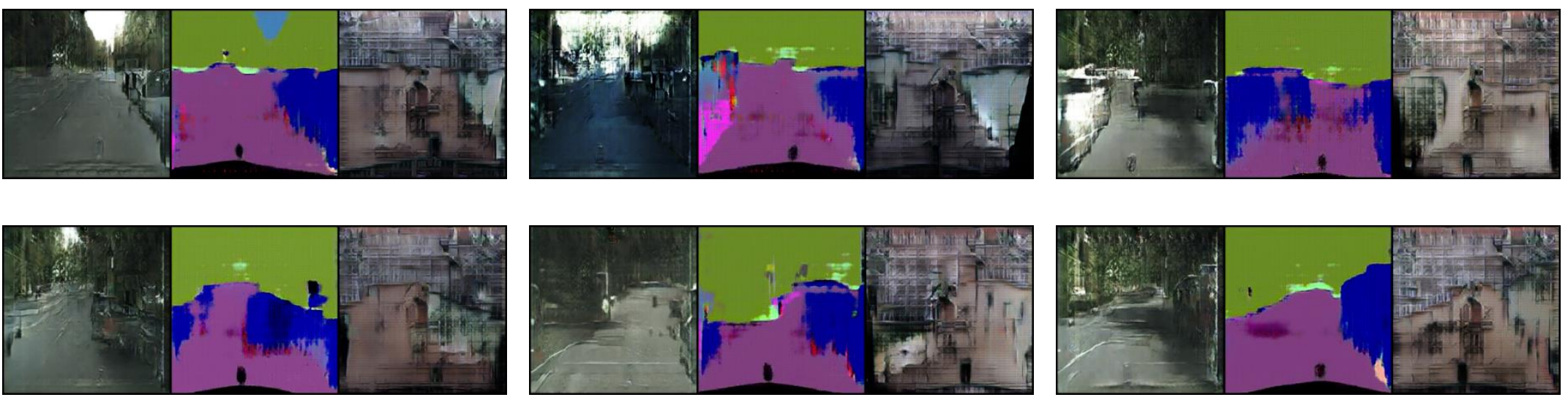}
	\vspace{-2mm}
	\caption{\small{Generated paired samples from models trained on facades$\leftrightarrow$labels$\leftrightarrow$cityscapes. The images in the first row are sequentially generated from left to right while those in the second row are generated from right to left. Both generation starts from a noise vector.  }}\label{fig:three}
	\vspace{-3mm}
\end{figure*}

Figures~\ref{fig:three_mnist} and~\ref{fig:three} show the generated samples from models trained on three-domain images. The generated images in each tuple are highly correlated. Interestingly, in Figure~\ref{fig:three}, the synthesized labels strive to be consistent with both the generated street scene and facade photos.  

\begin{table}[t!]
	\centering
	\caption{\small{Human evaluation results on the quality of generated pairs.}}\label{tab:human}
	\vspace{1mm}
	\begin{small}
		\begin{tabular}{c|cc}
			\toprule
			Method & Realism & Relevance \\
			\midrule
			\emph{Trained with paired data} & \\
			\midrule
			WGAN-GP + Pix2pix wins & 2.32\% &  3.1\% \\
			JointGAN wins &  17.93\% &36.32\% \\
			Not distinguishable & 79.75\% &60.58\% \\
			\midrule
			\emph{Trained with unpaired data} & \\
			\midrule
			WGAN-GP + CycleGAN wins & 0.13\% &  1.31\% \\
			JointGAN wins &  81.55\% &40.87\% \\
			Not distinguishable & 18.32\% &57.82\% \\
			\bottomrule
		\end{tabular}
	\end{small}
	\vspace{-4mm}
\end{table}

\subsubsection{Quantitative Results}
We perform a detailed quantitative analysis on the two-domain image-pair task. 

\vspace{-3mm}
\paragraph{Human Evaluation}
We perform human evaluation using Amazon Mechanical Turk (AMT), and present human evaluation results on the relevance and realism of generated pairs in both the cases with or without access to paired empirical samples. In each survey, we compare JointGAN and the two-step baseline by taking a random sample of 100 generated image pairs (5 datasets, 20 samples on each dataset), and ask the human evaluator to select which sample is more realistic and the content of which pairs are more relevant. 
We obtained roughly 44 responses per data sample (4378 samples in total) and the results are shown in Table~\ref{tab:human}. Clearly, human analysis suggest that our JointGAN produces higher-quality samples when compared with the two-step baseline, verifying the effectiveness of learning the marginal and conditional simultaneously.

\vspace{-2mm}
\paragraph{Relevance Score}
\begin{table*}[t!]
	\centering
	\vspace{-1mm}
	\caption{\small{Relevance scores of the generated pairs on the five two-domain image datasets.}}\label{tab:relevance_score}
	\begin{small}
		\begin{tabular}{c|ccccc}
			\toprule
			& edges$\leftrightarrow$shoes & edges$\leftrightarrow$handbags & labels$\leftrightarrow$cityscapes & labels$\leftrightarrow$facades & maps$\leftrightarrow$satellites \\
			\midrule
			True pairs & 0.684 & 0.672 & 0.591 & 0.529 & 0.514 \\
			Random pairs & 0.008 & 0.005 & 0.012 & 0.011 & 0.054 \\
			Other pairs & 0.113 & 0.139 & 0.092 & 0.076 & 0.081 \\
			\midrule
			WGAN-GP + Pix2pix & 0.352 & 0.343 & 0.301 & 0.288 & 0.125\\
			JointGAN (paired) &  0.488 & 0.489 & 0.377 & 0.364 & 0.328 \\
			\midrule
			WGAN-GP + CycleGAN & 0.203 & 0.195 & 0.201 & 0.139 & 0.091 \\
			JointGAN (unpaired) &  0.452 & 0.461 & 0.339 & 0.341 & 0.299 \\
			\bottomrule
		\end{tabular}
	\end{small}
\end{table*}

We use relevance score to evaluate the quality and relevance of two generated images. The relevance score is calculated as the cosine similarity between two images that are embedded into a shared latent space, which are learned via training a ranking model~\cite{huang2013learning}. Details are provided in Appendix B. The final relevance score is the average over all the individual relevance scores on each pair.
Results are summarized in Table~\ref{tab:relevance_score}. Our JointGAN provides significantly better results than the two-step baselines, especially when we do not have access to the paired empirical samples. 

Besides the results of our model and baselines, we also present results on three types of real images: ($i$) {\em True pairs:} this is the real image pairs from the same dataset but not used for training the ranking model; ($ii$) {\em  Random pairs}: the images are from the same dataset but the content of two images are not correlated; ($iii$) {\em  Other pairs}: the images are correlated but sampled from a dataset different from the training set. We can see in Table~\ref{tab:relevance_score} that the first one obtains a high relevance score while the latter two have a very low score, which shows that the relevance score metric assigns a low value when either the content of generated image pairs is not correlated or the images are not plausibly like the training set. It demonstrates that this metric correlates well with the quality of generated image pairs. 

\subsection{Joint modeling caption features and images}

\begin{figure}[t!]
	\centering
	\includegraphics[width=.48\textwidth]{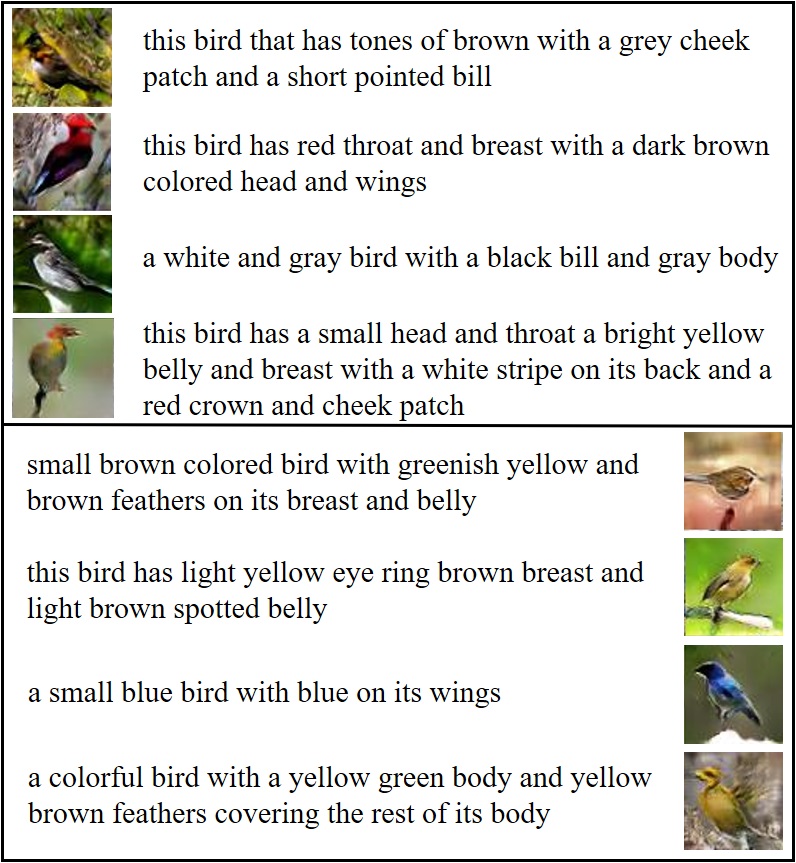}
	\vspace{-5mm}
	\caption{\small Generated paired samples of caption features and images (all data synthesized). Top block: from generated images to caption features. Bottom block: from generated caption features to images.}
	\vspace{-5mm}
	\label{fig:image-caption}
\end{figure}

\paragraph{Setup}
Our model is next evaluated on the Caltech-UCSD Birds dataset \cite{cub}, in which each image of bird is paired with 10 different captions. Since generating realistic text using GAN itself is a challenging task, in this work, we train our model on pairs of caption {\em features} and images. The caption features are obtained from a pretrained word-level CNN-LSTM autoencoder~\cite{gan2017learning}, which aims to achieve a one-to-one mapping between the captions and the features. We then train JointGAN based on the caption features and their corresponding images (the paired data for training JointGAN use CNN-generated text features, which avoids issues of training GAN for text generation). Finally to visualize the results, we use the pretrained LSTM decoder to decode the generated features back to captions.
We employ StackGAN-stage-I~\cite{zhang2017stackgan} for generating images from caption features while a CNN is utilized to generate caption features from images. Details are provided in Appendix D. 

\paragraph{Qualitative Results}
Figure \ref{fig:image-caption} shows the qualitative results of JointGAN: ($i$) generate images from noise and then conditionally generate caption features, and ($ii$) generate caption features from noise and then conditionally generate images. 
The results show high-quality and diverse image generation, 
and strong coherent relationship between each pair of the caption feature and image. 
It demonstrates the robustness of our model, in that it not only generates realistic multi-domain images but also handles well different datasets such as caption feature and image pairs.

\section{Conclusion}
%
We propose JointGAN, a new framework for multi-domain joint distribution learning. The joint distribution is learned via decomposing it into the product of a marginal and a conditional distribution(s), each learned via adversarial training. 
JointGAN allows interesting applications since it provides freedom to draw samples from various marginalized or conditional distributions.
We consider joint analysis of two and three domains, and demonstrate that JointGAN achieves significantly better results than a two-step baseline model, both qualitatively and quantitatively.

\vspace{-1mm}
\section*{Acknowledgements}
This research was supported in part by DARPA, DOE, NIH, ONR and NSF.
\clearpage
\bibliography{icml2018}
\bibliographystyle{icml2018}

\end{document}

%% file: mlVecMat.tex
\usepackage[utf8]{inputenc} 
\usepackage[T1]{fontenc}    
\usepackage{hyperref}       
\usepackage{url}            
\usepackage{booktabs}       
\usepackage{amsfonts}       
\usepackage{nicefrac}       
\usepackage{microtype}      

\usepackage{amsmath}
\usepackage{amsfonts}
\usepackage{amssymb}
\usepackage{wrapfig}
\usepackage{bm}
\usepackage{psfrag}
\usepackage{tikz}
\usepackage{epstopdf}

\usepackage{color}
\usepackage{tikz}
\usetikzlibrary{arrows,shapes,snakes,automata,backgrounds,fit,petri}
\usepackage{adjustbox}

\usepackage{algorithm}
\usepackage{algorithmic}

\newcommand{\eg}{\textit{e.g.}}
\newcommand{\ie}{\textit{i.e.}}

\newcommand{\beq}{\vspace{0mm}\begin{equation}}
\newcommand{\eeq}{\vspace{0mm}\end{equation}}
\newcommand{\beqs}{\vspace{0mm}\begin{eqnarray}}
\newcommand{\eeqs}{\vspace{0mm}\end{eqnarray}}
\newcommand{\barr}{\begin{array}}
\newcommand{\earr}{\end{array}}

\newcommand{\Imat}{{\bf I}}

\newcommand{\xv}{{\boldsymbol x}}
\newcommand{\yv}{{\boldsymbol y}}

\newcommand{\zv}{{\boldsymbol z}}

\newcommand{\alphav}{{\boldsymbol \alpha}}
\newcommand{\betav}{{\boldsymbol \beta}}
\newcommand{\gammav}{{\boldsymbol \gamma}}

\newcommand{\epsilonv}{{\boldsymbol \epsilon}}

\newcommand{\etav}{{\boldsymbol \eta}}
\newcommand{\thetav}{{\boldsymbol \theta}}

\newcommand{\nuv}{{\boldsymbol \nu}}

\newcommand{\phiv}{{\boldsymbol \phi}}

\newcommand{\psiv}{{\boldsymbol \psi}}
\newcommand{\omegav}{{\boldsymbol \omega}}

\newcommand{\E}{\mathbb{E}}

\newcommand{\zerov}{{\boldsymbol 0}}

\newcommand{\Xcal}{\mathcal{X}}
\newcommand{\Ycal}{\mathcal{Y}}

\newcommand{\Lcal}{\mathcal{L}}
\newcommand{\Ncal}{\mathcal{N}}

\newtheorem{prop}{Proposition}

%% file: JointGAN_camera.bbl
\begin{thebibliography}{39}
\providecommand{\natexlab}[1]{#1}
\providecommand{\url}[1]{\texttt{#1}}
\expandafter\ifx\csname urlstyle\endcsname\relax
  \providecommand{\doi}[1]{doi: #1}\else
  \providecommand{\doi}{doi: \begingroup \urlstyle{rm}\Url}\fi

\bibitem[Chen et~al.(2018)Chen, Dai, Pu, Zhou, Li, Su, Chen, and
  Carin]{pu2017symmetric}
Chen, L., Dai, S., Pu, Y., Zhou, E., Li, C., Su, Q., Chen, C., and Carin, L.
\newblock Symmetric variational autoencoder and connections to adversarial
  learning.
\newblock In \emph{AISTATS}, 2018.

\bibitem[Cordts et~al.(2016)Cordts, Omran, Ramos, Rehfeld, Enzweiler, Benenson,
  Franke, Roth, and Schiele]{cordts2016cityscapes}
Cordts, M., Omran, M., Ramos, S., Rehfeld, T., Enzweiler, M., Benenson, R.,
  Franke, U., Roth, S., and Schiele, B.
\newblock The cityscapes dataset for semantic urban scene understanding.
\newblock In \emph{CVPR}, 2016.

\bibitem[Donahue et~al.(2017)Donahue, Kr{\"a}henb{\"u}hl, and
  Darrell]{donahue2016adversarial}
Donahue, J., Kr{\"a}henb{\"u}hl, P., and Darrell, T.
\newblock Adversarial feature learning.
\newblock In \emph{ICLR}, 2017.

\bibitem[Dumoulin et~al.(2017)Dumoulin, Belghazi, Poole, Mastropietro, Lamb,
  Arjovsky, and Courville]{dumoulin2016adversarially}
Dumoulin, V., Belghazi, I., Poole, B., Mastropietro, O., Lamb, A., Arjovsky,
  M., and Courville, A.
\newblock Adversarially learned inference.
\newblock In \emph{ICLR}, 2017.

\bibitem[Gan et~al.(2017{\natexlab{a}})Gan, Chen, Wang, Pu, Zhang, Liu, Li, and
  Carin]{gan2017triangle}
Gan, Z., Chen, L., Wang, W., Pu, Y., Zhang, Y., Liu, H., Li, C., and Carin, L.
\newblock Triangle generative adversarial networks.
\newblock In \emph{NIPS}, 2017{\natexlab{a}}.

\bibitem[Gan et~al.(2017{\natexlab{b}})Gan, Pu, Henao, Li, He, and
  Carin]{gan2017learning}
Gan, Z., Pu, Y., Henao, R., Li, C., He, X., and Carin, L.
\newblock Learning generic sentence representations using convolutional neural
  networks.
\newblock In \emph{EMNLP}, 2017{\natexlab{b}}.

\bibitem[Goodfellow et~al.(2014)Goodfellow, Pouget-Abadie, Mirza, Xu,
  Warde-Farley, Ozair, Courville, and Bengio]{goodfellow2014generative}
Goodfellow, I., Pouget-Abadie, J., Mirza, M., Xu, B., Warde-Farley, D., Ozair,
  S., Courville, A., and Bengio, Y.
\newblock Generative adversarial nets.
\newblock In \emph{NIPS}, 2014.

\bibitem[Gulrajani et~al.(2017)Gulrajani, Ahmed, Arjovsky, Dumoulin, and
  Courville]{iwgan}
Gulrajani, I., Ahmed, F., Arjovsky, M., Dumoulin, V., and Courville, A.~C.
\newblock Improved training of {W}asserstein {GAN}s.
\newblock In \emph{NIPS}, 2017.

\bibitem[Hu et~al.(2017)Hu, Yang, Salakhutdinov, and Xing]{hu2017unifying}
Hu, Z., Yang, Z., Salakhutdinov, R., and Xing, E.~P.
\newblock On unifying deep generative models.
\newblock \emph{arXiv preprint arXiv:1706.00550}, 2017.

\bibitem[Huang et~al.(2013)Huang, He, Gao, Deng, Acero, and
  Heck]{huang2013learning}
Huang, P.-S., He, X., Gao, J., Deng, L., Acero, A., and Heck, L.
\newblock Learning deep structured semantic models for web search using
  clickthrough data.
\newblock In \emph{CIKM}, 2013.

\bibitem[Isola et~al.(2017)Isola, Zhu, Zhou, and Efros]{isola2016image}
Isola, P., Zhu, J.-Y., Zhou, T., and Efros, A.~A.
\newblock Image-to-image translation with conditional adversarial networks.
\newblock In \emph{CVPR}, 2017.

\bibitem[Kim et~al.(2017)Kim, Cha, Kim, Lee, and Kim]{kim2017learning}
Kim, T., Cha, M., Kim, H., Lee, J., and Kim, J.
\newblock Learning to discover cross-domain relations with generative
  adversarial networks.
\newblock In \emph{ICML}, 2017.

\bibitem[Kingma \& Ba(2014)Kingma and Ba]{kingma2014adam}
Kingma, D.~P. and Ba, J.
\newblock Adam: A method for stochastic optimization.
\newblock \emph{arXiv preprint arXiv:1412.6980}, 2014.

\bibitem[Kingma \& Welling(2013)Kingma and Welling]{kingma2013auto}
Kingma, D.~P. and Welling, M.
\newblock Auto-encoding variational bayes.
\newblock \emph{arXiv preprint arXiv:1312.6114}, 2013.

\bibitem[Larsen et~al.(2015)Larsen, S{\o}nderby, Larochelle, and
  Winther]{larsen2015autoencoding}
Larsen, A. B.~L., S{\o}nderby, S.~K., Larochelle, H., and Winther, O.
\newblock Autoencoding beyond pixels using a learned similarity metric.
\newblock \emph{arXiv preprint arXiv:1512.09300}, 2015.

\bibitem[Li et~al.(2017{\natexlab{a}})Li, Liu, Chen, Pu, Chen, Henao, and
  Carin]{li2017alice}
Li, C., Liu, H., Chen, C., Pu, Y., Chen, L., Henao, R., and Carin, L.
\newblock Alice: Towards understanding adversarial learning for joint
  distribution matching.
\newblock In \emph{NIPS}, 2017{\natexlab{a}}.

\bibitem[Li et~al.(2017{\natexlab{b}})Li, Xu, Zhu, and
  Zhang]{chongxuan2017triple}
Li, C., Xu, K., Zhu, J., and Zhang, B.
\newblock Triple generative adversarial nets.
\newblock In \emph{NIPS}, 2017{\natexlab{b}}.

\bibitem[Liu \& Tuzel(2016)Liu and Tuzel]{liu2016coupled}
Liu, M.-Y. and Tuzel, O.
\newblock Coupled generative adversarial networks.
\newblock In \emph{NIPS}, 2016.

\bibitem[Liu et~al.(2017)Liu, Breuel, and Kautz]{liu2017unsupervised}
Liu, M.-Y., Breuel, T., and Kautz, J.
\newblock Unsupervised image-to-image translation networks.
\newblock In \emph{NIPS}, 2017.

\bibitem[Makhzani et~al.(2015)Makhzani, Shlens, Jaitly, Goodfellow, and
  Frey]{makhzani2015adversarial}
Makhzani, A., Shlens, J., Jaitly, N., Goodfellow, I., and Frey, B.
\newblock Adversarial autoencoders.
\newblock \emph{arXiv preprint arXiv:1511.05644}, 2015.

\bibitem[Mescheder et~al.(2017)Mescheder, Nowozin, and
  Geiger]{mescheder2017adversarial}
Mescheder, L., Nowozin, S., and Geiger, A.
\newblock Adversarial variational bayes: Unifying variational autoencoders and
  generative adversarial networks.
\newblock In \emph{ICML}, 2017.

\bibitem[Mirza \& Osindero(2014)Mirza and Osindero]{mirza2014conditional}
Mirza, M. and Osindero, S.
\newblock Conditional generative adversarial nets.
\newblock In \emph{arXiv preprint arXiv:1411.1784}, 2014.

\bibitem[Perarnau et~al.(2016)Perarnau, van~de Weijer, Raducanu, and
  {\'A}lvarez]{perarnau2016invertible}
Perarnau, G., van~de Weijer, J., Raducanu, B., and {\'A}lvarez, J.~M.
\newblock Invertible conditional gans for image editing.
\newblock \emph{arXiv preprint arXiv:1611.06355}, 2016.

\bibitem[Pu et~al.(2016)Pu, Gan, Henao, Yuan, Li, Stevens, and
  Carin]{pu2016variational}
Pu, Y., Gan, Z., Henao, R., Yuan, X., Li, C., Stevens, A., and Carin, L.
\newblock Variational autoencoder for deep learning of images, labels and
  captions.
\newblock In \emph{NIPS}, 2016.

\bibitem[Pu et~al.(2017)Pu, Wang, Henao, Chen, Gan, Li, and
  Carin]{pu2017adversarial}
Pu, Y., Wang, W., Henao, R., Chen, L., Gan, Z., Li, C., and Carin, L.
\newblock Adversarial symmetric variational autoencoder.
\newblock In \emph{NIPS}, 2017.

\bibitem[Radford et~al.(2016)Radford, Metz, and
  Chintala]{radford2015unsupervised}
Radford, A., Metz, L., and Chintala, S.
\newblock Unsupervised representation learning with deep convolutional
  generative adversarial networks.
\newblock \emph{ICLR}, 2016.

\bibitem[Reed et~al.(2016)Reed, Akata, Yan, Logeswaran, Schiele, and
  Lee]{reed2016generative}
Reed, S., Akata, Z., Yan, X., Logeswaran, L., Schiele, B., and Lee, H.
\newblock Generative adversarial text to image synthesis.
\newblock In \emph{ICML}, 2016.

\bibitem[Ronneberger et~al.(2015)Ronneberger, Fischer, and Brox]{Unet}
Ronneberger, O., Fischer, P., and Brox, T.
\newblock U-net: Convolutional networks for biomedical image segmentation.
\newblock In \emph{MICCAI}, 2015.

\bibitem[Tao et~al.(2018)Tao, Chen, Henao, Feng, and Carin]{Tao2018chi2}
Tao, C., Chen, L., Henao, R., Feng, J., and Carin, L.
\newblock Chi-square generative adversarial network.
\newblock In \emph{ICML}, 2018.

\bibitem[Tyle{\v{c}}ek \& {\v{S}}{\'a}ra(2013)Tyle{\v{c}}ek and
  {\v{S}}{\'a}ra]{tylevcek2013spatial}
Tyle{\v{c}}ek, R. and {\v{S}}{\'a}ra, R.
\newblock Spatial pattern templates for recognition of objects with regular
  structure.
\newblock In \emph{GCPR}, 2013.

\bibitem[Welinder et~al.(2010)Welinder, Branson, Mita, Wah, Schroff, Belongie,
  and Perona]{cub}
Welinder, P., Branson, S., Mita, T., Wah, C., Schroff, F., Belongie, S., and
  Perona, P.
\newblock {Caltech-UCSD Birds 200}.
\newblock Technical report, California Institute of Technology, 2010.

\bibitem[Xu et~al.(2017)Xu, Zhang, Huang, Zhang, Gan, Huang, and
  He]{xu2017attngan}
Xu, T., Zhang, P., Huang, Q., Zhang, H., Gan, Z., Huang, X., and He, X.
\newblock Attngan: Fine-grained text to image generation with attentional
  generative adversarial networks.
\newblock \emph{arXiv preprint arXiv:1711.10485}, 2017.

\bibitem[Yi et~al.(2017)Yi, Zhang, Tan, and Gong]{yi2017dualgan}
Yi, Z., Zhang, H., Tan, P., and Gong, M.
\newblock Dualgan: Unsupervised dual learning for image-to-image translation.
\newblock In \emph{CVPR}, 2017.

\bibitem[Yu \& Grauman(2014)Yu and Grauman]{yu2014fine}
Yu, A. and Grauman, K.
\newblock Fine-grained visual comparisons with local learning.
\newblock In \emph{CVPR}, 2014.

\bibitem[Yu et~al.(2017)Yu, Zhang, Wang, and Yu]{yu2017seqgan}
Yu, L., Zhang, W., Wang, J., and Yu, Y.
\newblock Seqgan: Sequence generative adversarial nets with policy gradient.
\newblock In \emph{AAAI}, 2017.

\bibitem[Zhang et~al.(2017{\natexlab{a}})Zhang, Xu, Li, Zhang, and
  Metaxas]{zhang2017stackgan}
Zhang, H., Xu, T., Li, H., Zhang, S., and Metaxas, D.
\newblock Stackgan: Text to photo-realistic image synthesis with stacked
  generative adversarial networks.
\newblock In \emph{ICCV}, 2017{\natexlab{a}}.

\bibitem[Zhang et~al.(2017{\natexlab{b}})Zhang, Gan, Fan, Chen, Henao, Shen,
  and Carin]{zhang2017adversarial}
Zhang, Y., Gan, Z., Fan, K., Chen, Z., Henao, R., Shen, D., and Carin, L.
\newblock Adversarial feature matching for text generation.
\newblock In \emph{ICML}, 2017{\natexlab{b}}.

\bibitem[Zhu et~al.(2016)Zhu, Kr{\"a}henb{\"u}hl, Shechtman, and
  Efros]{zhu2016generative}
Zhu, J.-Y., Kr{\"a}henb{\"u}hl, P., Shechtman, E., and Efros, A.~A.
\newblock Generative visual manipulation on the natural image manifold.
\newblock In \emph{ECCV}, 2016.

\bibitem[Zhu et~al.(2017)Zhu, Park, Isola, and Efros]{zhu2017unpaired}
Zhu, J.-Y., Park, T., Isola, P., and Efros, A.~A.
\newblock Unpaired image-to-image translation using cycle-consistent
  adversarial networks.
\newblock In \emph{CVPR}, 2017.

\end{thebibliography}
